\pdfoutput=1

\documentclass[11pt]{article}
\DeclareUnicodeCharacter{FF1A}{:}

\usepackage[preprint]{acl}

\usepackage{times}
\usepackage{latexsym}
\usepackage{hyperref}       
\usepackage{url}            
\usepackage{booktabs}       
\usepackage{amsfonts}       
\usepackage{nicefrac}       
\usepackage{xcolor}         
\definecolor{grey}{rgb}{0.5, 0.5, 0.5}
\usepackage{mdframed}
\usepackage{threeparttable}
\usepackage{graphics}
\usepackage{multirow}
\usepackage{rotating}
\usepackage{array}
\usepackage{caption}
\usepackage{xcolor}
\usepackage{colortbl}
\usepackage{listings}
\usepackage{enumitem}
\usepackage{amsmath}
\usepackage{cleveref}
\usepackage{hyperref}
\usepackage{xspace}
\usepackage{float}

\usepackage{natbib}
\setcitestyle{authoryear}

\usepackage[T1]{fontenc}

\usepackage[utf8]{inputenc}

\usepackage{microtype}

\usepackage{inconsolata}

\usepackage{graphicx}

\title{GraphKV: Breaking the Static Selection Paradigm with\\ Graph-Based KV Cache Eviction}
\author{
  Xuelin Li\textsuperscript{1} \
  Xiangqi Jin\textsuperscript{1} \
  Linfeng Zhang\textsuperscript{1,†}
  \\
  \\
  \textsuperscript{1}EPIC Lab, Shanghai Jiao Tong University \\
  \normalfont\texttt{lxl.cooper@outlook.com, zhanglinfeng@sjtu.edu.cn} 
}

\usepackage{graphicx} 
\usepackage{amsmath}   
\usepackage{amssymb}
\usepackage{array}
\usepackage{float} 
\usepackage{booktabs}
\usepackage{multirow}  
\usepackage{rotating}
\usepackage{makecell}
\usepackage{algorithm}
\usepackage{algpseudocode}
\usepackage{tikz}

\begin{document}
\maketitle

\begin{abstract}
\footnotetext{†Corresponding author}
Efficient Key-Value (KV) cache management is essential for processing long text sequences in large language models (LLMs), where memory constraints often limit performance. Conventional KV eviction strategies, such as top-k selection based on attention scores, depend on static heuristics that fail to capture the evolving implicit dependencies among tokens during inference. To overcome this, we propose GraphKV, a graph-based framework that redefines token selection for KV cache compression. In GraphKV, \emph{\textbf{tokens}} are modeled as \emph{\textbf{nodes}} with importance scores, and \emph{\textbf{edges}} represent their \emph{\textbf{similarity relationships}}. Through a decay-signal-propagation mechanism, token importance is dynamically updated by propagating information across the graph, enabling adaptive retention of the most contextually significant tokens. 
GraphKV can be seamlessly utilized in existing KV cache eviction methods such as SnapKV and PyramidKV in a plug-and-play manner.
\emph{Codes will be released on Github}.

\end{abstract}

\section{Introduction}

Large language models (LLMs) have enhanced proficiency in processing long-text, improving performance in multi-turn dialogues \cite{chiang2023vicuna}, document summarization \cite{zhang2024benchmarking}, question answering \cite{bai2023longbench}, information retrieval \cite{zhu2023large}, and code generation \cite{li2025structured}. New models such as GPT-4 \cite{achiam2023gpt}, Claude 3.5 \cite{anthropic2024claude}, LLaMA 3.1 \cite{grattafiori2024llama}, and Mistral Large 2 have extended token processing capacities beyond 128K. In the context of long-text processing, key value caching (KV cache) \cite{yang2024pyramidinfer,wu2024scope} is a crucial technique for improving the efficiency of LLMs. As the context length grows, the size of the KV cache increases linearly, leading to significant memory and computational overhead. \citet{liu2024deepseek}; \citet{singhania2024loki} partially address this issue by leveraging low-rank decomposition to approximate the full-rank KV cache during the training phase. However, efficiently optimizing the key-value cache without extra training is vital for inference on long contexts under memory limitations, especially in standard deployment scenarios with a fixed model architecture.
\begin{figure}[t!] 
  \centering
  \includegraphics[width=0.48\textwidth]{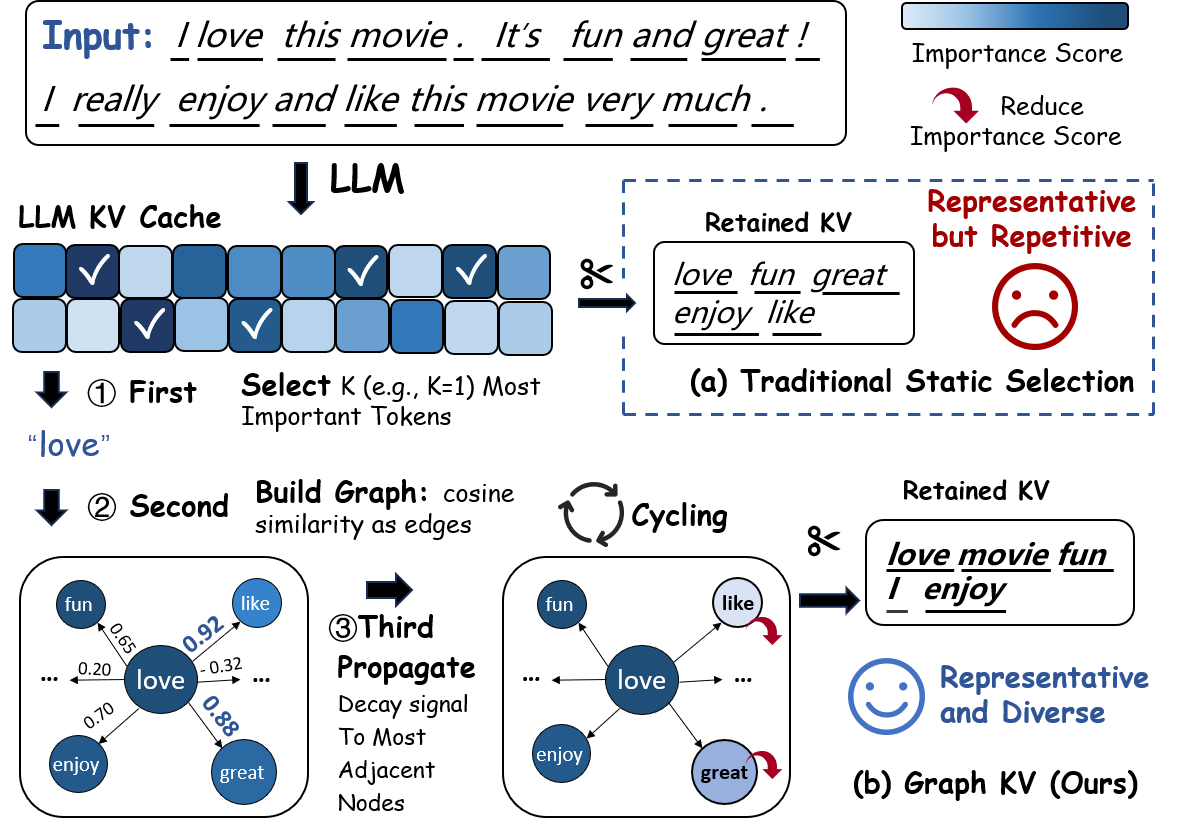} 
  \caption{\textbf{Overview of GraphKV.} Tokens are defined as nodes with their initial importance scores. 
  The cosine similarity between tokens is defined as the edges.
  GraphKV aims to firstly identify \emph{K} most important nodes (\emph{e.g., ``love'', \emph{K=1}}), and then propagate \textbf{decay signal} to their most adjacent tokens (\emph{e.g.,} ``like'' and ``great'') to obtain diverse retained KV. This propagation can be performed one or multiple times.} 
  \label{head1}    
\end{figure}

Building on training-free approaches to optimize the key-value cache, token eviction has emerged as a viable and effective method to compress the KV cache by selectively removing less important tokens. The core challenge in token eviction lies in accurately identifying and retaining the most critical tokens while discarding redundant ones, without compromising the model's performance.

\begin{figure*}[t!] 
  \centering
  \includegraphics[width=1\textwidth]{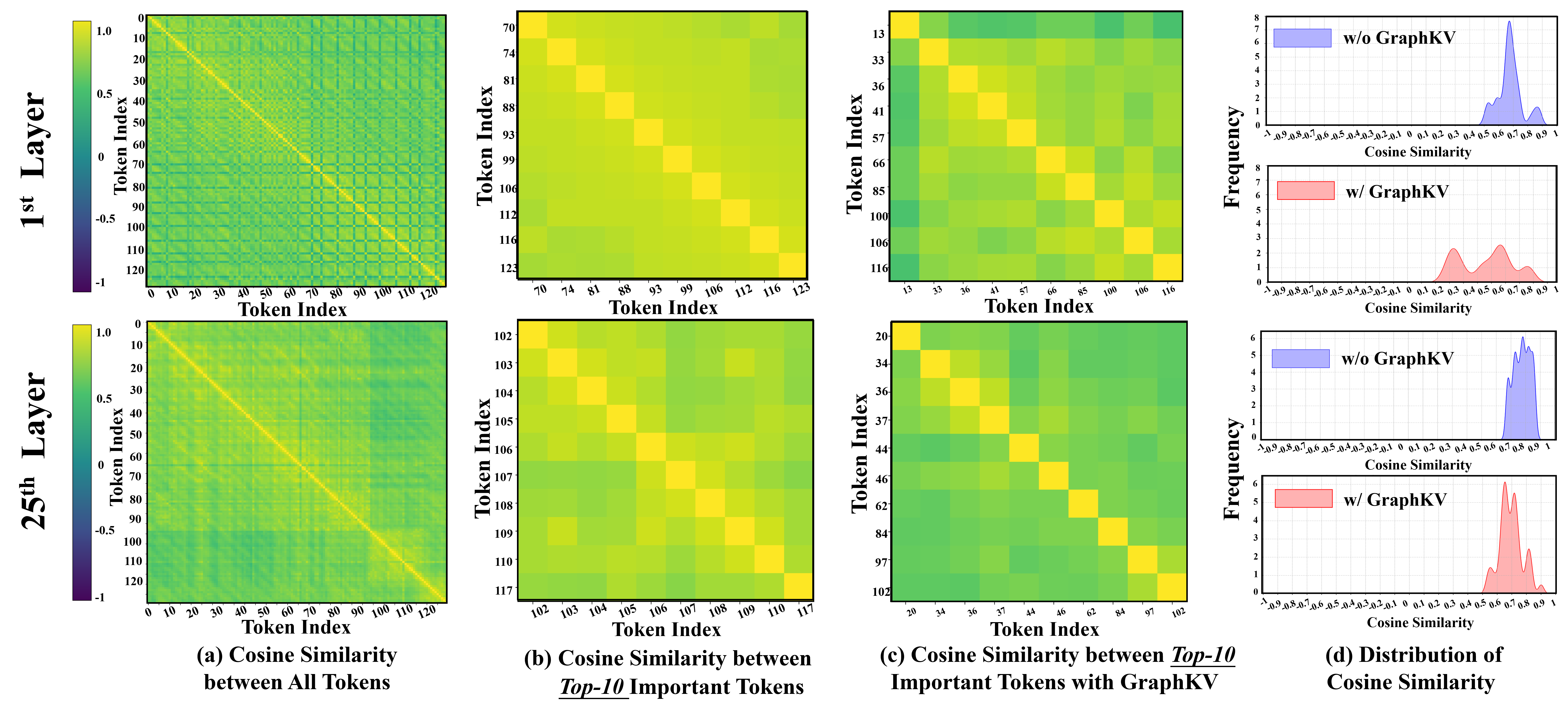}
  \vspace{-1cm}
  \caption{Cosine similarity of keys and the frequency distribution of cosine similarity, measured using Llama3-8B on a sample from the HotpotQA dataset in LongBench, with truncated 128 tokens for visualization. \textbf{(a)} illustrates the cosine similarity across all 128 tokens. \textbf{(b)} depicts the cosine similarity for the top-10 tokens with highest importance scores. \textbf{(c)} presents the cosine similarity for the top-10 tokens after GraphKV's decay signal propagation. \textbf{(d)} compares the frequency distribution of cosine similarity for the keys before (w/o GraphKV) and after (w/ GraphKV) GraphKV's decay signal propagation.
  }
  \label{insight}
\end{figure*}

Most existing token eviction strategies~\cite{zhang2023h2o, li2024snapkv, cai2024pyramidkv,guo2024attention} reduce KV cache size via identifying important tokens based on their importance scores (\emph{e.g.,} attention scores).
Then, as shown in Figure~\ref{head1}, a \emph{static} selection is utilized to remove the KV cache of the redundant tokens and only retain the most important ones.
Such a one-stage selection can identify the most representative tokens (\emph{i.e., } tokens which are most relevant to the following query), but always suffers from the problem of \emph{duplication}, since the important tokens usually contains similar semantics, as shown in Figure~\ref{insight} and will be discussed in Section~\ref{sec:observation}. This observation highlights that 
retaining multiple tokens with high importance can lead to redundancy if they are highly similar, while the other less important but dissimilar tokens that contain more diverse semantic information can contribute uniquely to the overall context.  

To bridge the gap, as shown in Figure~\ref{head1}, we propose GraphKV, which formulates KV cache as a graph, employing a decay-signal-propagation mechanism in the graph for importance propagation among tokens, enabling dynamic KV selection. 
Inspired by graph-based methods, where dynamic node updating mechanisms iteratively refine node states by aggregating information from neighbors, we construct a graph where tokens are defined as nodes, and the similarities among tokens are represented as edges. The importance scores defined by any previous KV cache eviction methods
serve as the initial value for each node. Concretely, in the first step, we select only a few tokens (\emph{K}) with the highest importance scores and compute their cosine similarity with other tokens to initialize the graph. In the second step, we perform a \textbf{decay-signal-propagation} over the graph, reducing the importance scores of the tokens that are most adjacent (\emph{i.e.,} adjacent) to the previously selected important nodes. Such a negative propagation helps to reduce the possibility of retaining multiple highly similar tokens, and it can be performed by one all multiple times. Finally, we remove the KV cache of tokens with lower scores after the propagation. Based on GraphKV, the retained KV cache can be both representative and diverse, minimizing the information loss from KV eviction.

Notably, GraphKV is not a new KV eviction method that introduces a new important score, it is a framework that can be directly applied to any previous KV cache eviction methods in a plug-and-play manner. The only additional computation in GraphKV is to compute the cosine similarity between the selected \emph{K} most important tokens and other tokens. Since we define $K << N$, such a computation shows linear complexity as $O(N\times K) \approx O(N)$. which is ignorable. Experimental results validate GraphKV’s effectiveness. For example, on the LongBench QA task, GraphKV outperforms the suboptimal Knorm \cite{devoto2024simple} method by 45.88\% and achieves approximately 3\% improvement over state-of-the-art models SnapKV \cite{li2024snapkv} and PyramidKV \cite{cai2024pyramidkv} with a KV cache size of 512 on the LLaMA-8B model. Additionally, GraphKV demonstrates superior performance on the Needle in a Haystack benchmark, further highlighting its ability to retain critical context details in long-context scenarios.

In summary, our contributions are as follows:
\begin{itemize}
    \item Formulation of the KV cache as a graph, with tokens as nodes and semantic similarities as edges, enables dynamic modeling of token relationships for efficient eviction. 
\end{itemize}
\begin{itemize}
    \item A decay-signal-propagation mechanism in GraphKV iteratively propagates decay information to suppress redundant tokens, leveraging token similarity to prioritize diverse and representative context. 
\end{itemize}
\begin{itemize}
    \item Extensive experiments on LongBench and Needle-in-a-Haystack benchmarks demonstrate that GraphKV can be applied to any previous KV cache eviction methods in a plug-and-play manner with significant accuracy improvements under the same compression ratio.
\end{itemize}
\section{Related works}

\noindent \textbf{KV Cache Eviction.} With KV cache size scaling linearly with sequence length, efficient management of KV cache has garnered significant attention as an effective way to mitigate the memory constraints of LLMs when processing long contexts. Numerous studies have proposed token eviction strategies to reduce memory overhead while maintaining inference performance by selectively retaining only the most relevant tokens in the cache.
Heavy Hitter Oracle (H2O)~\cite{zhang2023h2o} introduces a dynamic eviction policy that balances retention of recent and historically significant tokens, optimizing memory usage without sacrificing critical information. Similarly, SnapKV~\cite{li2024snapkv} enhances efficiency by clustering significant KV positions based on attention scores of an observation window, while PyramidKV~\cite{cai2024pyramidkv} adopts a layer-wise approach, allocating variable KV budgets according to each layer’s attention demands. StreamingLLM~\cite{streamingllm} enables models trained on finite attention windows to process infinite sequences without retraining by preserving initial attention sinks and recent local tokens. FastGen~\cite{fastgen} employs an adaptive strategy, tailoring KV retention to the behavior of individual attention heads. Despite their success in reducing cache size, these methods predominantly rely on static importance scores, overlooking the dynamic, implicit relationships among tokens. This limitation can lead to suboptimal retention decisions, motivating the need for approaches that capture token interdependencies more effectively.

\begin{figure*}[t!]
  \includegraphics[width=1\textwidth]{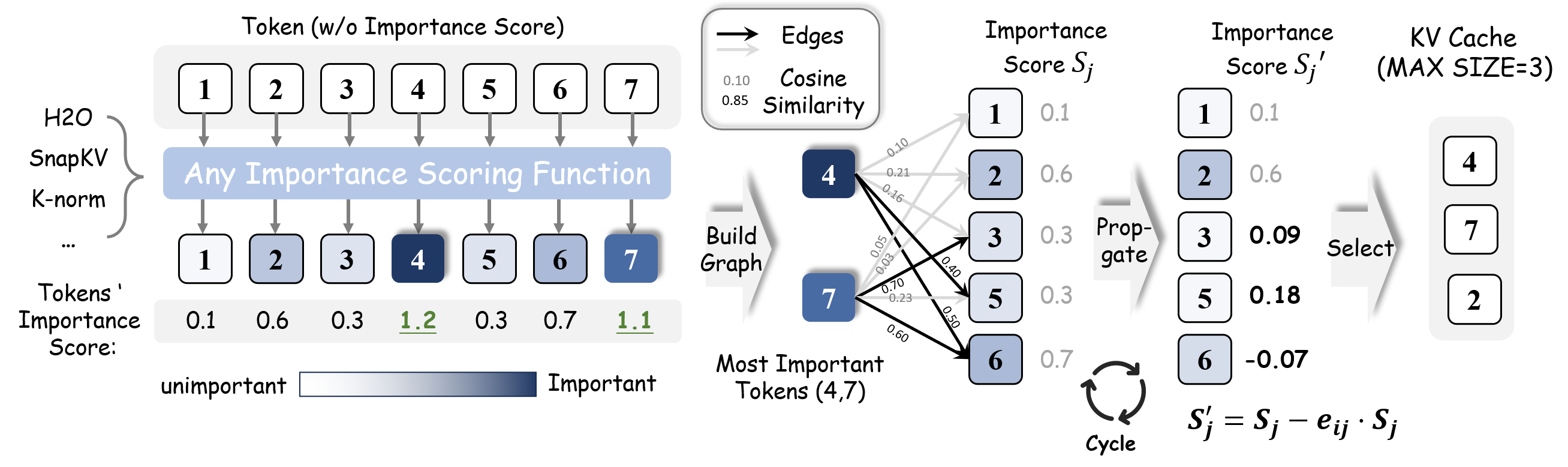} 
  \vspace{-0.4cm}
  \caption{The pipeline of GraphKV. (I) \textbf{Initiate Importance Score}: We initialize importance scores for each token with any importance scoring function. (II) \textbf{Build Graph}: We select tokens with the highest importance scores as source nodes and compute the key cosine similarity between source nodes and other nodes, retaining the highest similarity values as edges in the graph. (III) \textbf{Decay Signal Propagation}: After constructing the graph, we perform multiple rounds of decay signal propagation on nodes adjacent to source nodes, attenuating their importance scores based on edge weights to reduce semantic redundancy. (IV) \textbf{Select Important Tokens}: Based on the updated scores, we reselect the most important tokens to balance significance and diversity for the compressed KV cache.}
  \label{exp1}    
\end{figure*}
\section{Observation}
\label{sec:observation}
 Most static-importance-score-driven methods utilize Top-K to select tokens based on their importance. However, \citet{chen2024magicpig} notes that Top-K's reliance on attention score magnitude for importance estimation is biased and fails to capture the true distribution of important tokens. To better understand the limitations of static-importance-score-driven token eviction, we further analyzed the interplay between token importance and semantic similarity in the KV cache of LLaMA-8B, using a sample from the HotpotQA dataset in LongBench. Figure \ref{insight} visualizes the cosine similarity of key states and the frequency distribution of cosine similarity values for 128 truncated tokens, comparing the effects of GraphKV's decay signal propagation across two layers. Subfigures (a) and (b) show high cosine similarity among top-10 tokens by importance, with (b) having a brighter heatmap. Subfigure (d) reveals high mean and low variance in cosine similarity before GraphKV's decay signal propagation, indicating potential redundancy among high-importance tokens. This aligns with \cite{wang2024model}. Additionally, we identified several tokens with moderate importance scores but low key similarity, likely encoding critical semantic diversity. We extended our analysis across different model layers to verify the generalizability of this finding, and results show the phenomenon persists across layers.

This observation underscores a key insight: \textit{\textbf{while high-scoring tokens are important, their similarity often leads to redundancy, whereas less similar tokens with moderate scores contribute unique semantic diversity.}} In token eviction for KV cache compression, prioritizing these diverse, non-redundant tokens is crucial for preserving critical context details under memory constraints. Therefore, our proposed GraphKV leverages this insight by dynamically refining importance scores through similarity-based decay signal propagation, as demonstrated in Subfigures (c) and (d). Specifically, (c) shows that after GraphKV's decay propagation, the heatmap exhibits significantly darker colors, indicating a notable reduction in the key similarity among the top-10 tokens. Moreover, (d) shows the post-propagation distribution exhibits a lower mean and reduced variance compared to the pre-propagation distribution, further confirming the decline in key similarity distribution.

\section{Methodology}

\subsection{Problem Formulation}
Token eviction aims to retain a minimal subset of the KV cache while preserving as much contextual information as possible. Formally, we consider an LLM with \(M\) transformer layers processing a sequence of prompt tokens \(\mathbf{X}_{input} = [x_1,...,x_n]\), where \(n\) represents the total number of tokens in the input sequence. For the \(l\text{-}th\) layer \((l \in {0,1,..., M-1})\), the full key and value matrices are denoted as \(\mathbf{K}^{l}, \mathbf{V}^{l} \in \mathbb{R}^{n \times d}\), where \(d\) is the hidden dimension of the model. The objective of token eviction is to identify compressed sub-matrices \(\mathbf{K}_s^l, \mathbf{V}_s^l \in \mathbb{R}^{k_l \times d}\), where \(k_l \textless n\) is the layer-specific cache budget, while minimizing performance degradation. Formally, given a dataset \(D\) and a performance metric \(S_{core} (\cdot)\), the compressed model using \(\mathbf{K}_s^l\) and \(\mathbf{V}_s^l\) is expected to attain similar results with the full model: \(\mathrm{S_{core}}(\mathbf{K}_s^l, \mathbf{V}_s^l, \mathcal{D}) \approx \mathrm{S_{core}}(\mathbf{K}^l, \mathbf{V}^l, \mathcal{D})\).

\subsection{Sparse Graph Building}
Note that GraphKV is a graph-based framework designed to dynamically manage the Key-Value(KV) cache by modeling token relationships using a graph structure. First of all, we define the KV cache as a weighted graph \(G = (O, E) \). Specifically, each token \(x_i\) in the input sequence corresponds to a node \(o_i \in O \), with an initial importance score \(s_i\). Notably, this score can be initialized flexibly using important indicator of diverse established methods, such as attention-based scores from SnapKV, PyramidKV and CAKE, or KNorm, ensuring compatibility with prior high-performing frameworks. Inspired by the observation in \ref{sec:observation}, we model the cosine similarity between the keys of token pairs as an edge \(e_{ij} \in E\) in the weighted graph \(G\). This is formally expressed as followed:
 \begin{equation}
e_{ij} = \frac{\langle k_i, k_j \rangle}{\| k_i \| \| k_j \|}
\end{equation}
where \( k_i \) and \( k_j \) are key vectors of the token pairs \(x_i\) and \(x_j\) respectively. 

It is worth noting that computing the similarity for all node pairs in the graph incurs significant computational overhead, particularly when processing long input sequences with a large number of tokens. To address this, we further sparsify the graph. Since the graph contains numerous nodes with low importance scores, which we consider to be tokens with minimal semantic relevance, we isolate these low-scoring nodes by removing their edges. Conversely, we prioritize the similarity relationships among the more significant tokens. Therefore, we adopt a top-\( k \) selection strategy to identify the most important source nodes, as follows:
\begin{equation}
\begin{split}
O_{\text{source}} &= \{ o_i \mid s_i \text{ ranks in top-}k, \\
&\quad i \in \{1, \dots, n\} \}
\end{split}
\end{equation}
where \(k\) is a hyperparameter tied to the layer-specific cache budget \(k_l\). We compute the similarity only between these selected source nodes \(O_{\text{source}}\) and all other nodes, thereby completing the sparsification of the graph. Sparse graph structure building approach enables GraphKV to efficiently and effectively capture token relationships within the KV cache.

\subsection{Decay Signal Propagation}
Once the sparse graph \(G = (O, E)\) is constructed, we introduce a dynamic decay-signal-propagation mechanism to refine token importance scores and eliminate semantic redundancy caused by token similarity. This process leverages the graph structure to propagate redundant signal across nodes, ensuring that the retained tokens are both significant and contextually diverse. 

Initially, for each node \( o_i \in O_{\text{source}}\), we define its neighborhood as the set of nodes \( o_j \) whose edge weights are among the top-\( m \):
\begin{equation}
N(o_i) = \{ o_j \mid e_{ij} \geq e_{i(m)}, j \neq i \},
\end{equation}
where \( e_{i(m)} \) is the \( m \)-th largest edge weight among \( \{ e_{ij} \mid j \neq i \} \), \(m\) is a hyperparameter indicating the number of nodes in the neighborhood. This step ensures that decay score propagation targets tokens with high semantic overlap while preserving the diversity of retained tokens. 

Furthermore, to prevent the over-representation of similar tokens, we apply a decay function to the importance scores of nodes in \(N(O_i)\). The decay is proportional to their similarity to the source node, reducing the scores of redundant tokens. For a node \(o_j \in N(o_i)\), the decayed score after one round of propagation as followed :
\begin{equation}
s_j' = s_j - \cdot e_{ij} \cdot s_j,
\end{equation}
where  \(s_j'\) represents the refinded scores. 
To capture broader contextual dependencies, we extend this process over \( T \) rounds of propagation (e.g., \( T = 3 \)), where in each round \( t \), the updated score of a node \( o_j \) is computed as:
\begin{equation}
s_j^{(t)} = s_j^{(t-1)} \cdot \prod_{o_i \in O_{\text{source}}, o_j \in N(o_i)} (1 - e_{ij})
\end{equation}
with \( s_j^{(0)} = s_j \), we aggregate cumulative decay from multiple source nodes to heavily suppress tokens similar to several retained ones. 
Finally, after \( T \) rounds, the updated scores \( s_j^{(T)} \) determine the final \( k_l \) tokens for the compressed KV cache sub-matrices \( \mathbf{K}_s^l \) and \( \mathbf{V}_s^l \), retaining those with the highest \( s_j^{(T)} \) to balance importance and diversity.

The decay-signal-propagation mechanism innovates over static-importance-score-driven methods by dynamically updating scores across multiple rounds, suppressing redundancy through similarity-based decay and offering flexibility by integrating initial scores from prior frameworks.

\begin{table*}[h]
\centering
\resizebox{\textwidth}{!}{
\begin{tabular}
{lccccccccccccccccc}

\specialrule{1pt}{0pt}{2pt}
\multirow{6}{*}{Method} & \multicolumn{6}{c}{Information Localization Task} & \multicolumn{10}{c}{Information Aggregation Task} & \multirow{5}{*}{Avg.} \\
\cmidrule(lr){2-7} \cmidrule(lr){8-17}
& \multicolumn{3}{c}{Single-Document QA} & \multicolumn{3}{c}{Multi-Document QA}& \multicolumn{3}{c}{Summarization}& \multicolumn{3}{c}{Few-shot Learning}& \multicolumn{2}{c}{Synthetic} & \multicolumn{2}{c}{Code} &  \\
\cmidrule(lr){2-4}\cmidrule(lr){5-7}\cmidrule(lr){8-10}\cmidrule(lr){11-13}\cmidrule(lr){14-15}\cmidrule(lr){16-17}
& \rotatebox[origin=c]{30}{NrtvQA} & \rotatebox[origin=c]{30}{Qasper} & \rotatebox[origin=c]{30}{MF-en} & \rotatebox[origin=c]{30}{HotpotQA} & \rotatebox[origin=c]{30}{2WikiMQA} & \rotatebox[origin=c]{30}{Musique} & \rotatebox[origin=c]{30}{GovReport} & \rotatebox[origin=c]{30}{QMSum} & \rotatebox[origin=c]{30}{MultiNews} & \rotatebox[origin=c]{30}{TREC} & \rotatebox[origin=c]{30}{TriviaQA} & \rotatebox[origin=c]{30}{SAMSum} & \rotatebox[origin=c]{30}{PCount} & \rotatebox[origin=c]{30}{PRe} & \rotatebox[origin=c]{30}{Lcc} & \rotatebox[origin=c]{30}{RB-P} & \\ 

\cmidrule(lr){2-17}
& F1 & F1 & F1 & F1 & F1 & F1 & R-L & R-L & R-L & Acc (CLS) & F1 & R-L & Acc (EM) & Acc (EM) & Edit Sim & Edit Sim \\

\midrule
\multicolumn{18}{c}{Llama2-7B-Chat, KV Size = 512} \\
\arrayrulecolor{black}\midrule
\textit{\textcolor{grey}{\textbf{Full}}} &
\textit{\textcolor{grey}{18.39}} &
\textit{\textcolor{grey}{20.10}} &
\textit{\textcolor{grey}{35.67}} &
\textit{\textcolor{grey}{31.25}} &
\textit{\textcolor{grey}{25.73}} &
\textit{\textcolor{grey}{10.64}} &
\textit{\textcolor{grey}{25.67}} &
\textit{\textcolor{grey}{20.89}} &
\textit{\textcolor{grey}{26.34}} &
\textit{\textcolor{grey}{64.00}} &
\textit{\textcolor{grey}{83.38}} &
\textit{\textcolor{grey}{40.90}} &
\textit{\textcolor{grey}{5.50}} &
\textit{\textcolor{grey}{10.00}}  &
\textit{\textcolor{grey}{58.67}} &
\textit{\textcolor{grey}{53.00}} &
\textit{\textcolor{grey}{33.13}} \\
\textbf{KNorm} & 6.72 & 8.10 & 7.94 & 8.38 & 7.06 & 2.72 & 16.60 & 15.69 & 19.69 & 19.50 & 27.26 & 11.86 & 4.50 & 1.67 & 28.60 & 22.13 & 13.03\\
\cellcolor{grey!10}~+ GraphKV & \cellcolor{grey!10}12.57 & \cellcolor{grey!10}15.55 & \cellcolor{grey!10}20.01 & \cellcolor{grey!10}26.15 & \cellcolor{grey!10}20.87 & \cellcolor{grey!10}6.52 & \cellcolor{grey!10}18.86 & \cellcolor{grey!10}18.82 & \cellcolor{grey!10}20.61 & \cellcolor{grey!10}31.50 & \cellcolor{grey!10}79.21 & \cellcolor{grey!10}33.43 & \cellcolor{grey!10}5.00 & \cellcolor{grey!10}8.00 & \cellcolor{grey!10}44.35 & \cellcolor{grey!10}41.67 & \cellcolor{grey!10}\textbf{25.20} \\
\textbf{SnapKV}& 16.01 & 19.11 & 32.40 & 32.25 & 24.18 & 10.47 & 19.96 & 20.97 & 23.50 & 62.00 &82.70 & 39.11 & 6.00 & 11.00 & 58.06 & 53.50 & 31.95\\
\cellcolor{grey!10}~+ GraphKV & \cellcolor{grey!10}16.39 & \cellcolor{grey!10}20.12 & \cellcolor{grey!10}33.58 & \cellcolor{grey!10}33.44 & \cellcolor{grey!10}25.19 & \cellcolor{grey!10}10.66 & \cellcolor{grey!10}20.23 & \cellcolor{grey!10}21.09 & \cellcolor{grey!10}23.38 & \cellcolor{grey!10}62.00 & \cellcolor{grey!10}83.21 & \cellcolor{grey!10}39.56 & \cellcolor{grey!10}6.00 & \cellcolor{grey!10}11.00 & \cellcolor{grey!10}57.65 & \cellcolor{grey!10}54.15 & \cellcolor{grey!10}\textbf{32.35} \\
\textbf{PyramidKV} & 17.22 & 19.82 & 34.15 & 32.29 & 24.41 & 10.08 & 20.28 & 20.47 & 23.46 & 62.50 & 83.10 & 38.61 & 6.00 & 11.50 & 57.17 & 51.62 & 32.04 \\
\cellcolor{grey!10}~+ GraphKV & \cellcolor{grey!10}16.39 & \cellcolor{grey!10}19.42 & \cellcolor{grey!10}35.62 & \cellcolor{grey!10}32.60 & \cellcolor{grey!10}25.85 & \cellcolor{grey!10}9.91 & \cellcolor{grey!10}20.43 & \cellcolor{grey!10}20.72 & \cellcolor{grey!10}23.64 & \cellcolor{grey!10}62.50 & \cellcolor{grey!10}83.43 & \cellcolor{grey!10}39.16 & \cellcolor{grey!10}6.00 & \cellcolor{grey!10}11.50 & \cellcolor{grey!10}57.11 & \cellcolor{grey!10}51.77 &  \cellcolor{grey!10}\textbf{32.25} \\

\arrayrulecolor{black}\midrule
\multicolumn{18}{c}{Llama3-8B-Instruct, KV Size = 512} \\
\arrayrulecolor{black}\midrule
\textit{\textcolor{grey}{\textbf{Full}}} &
\textit{\textcolor{grey}{25.56}} &
\textit{\textcolor{grey}{32.21}} &
\textit{\textcolor{grey}{39.65}} &
\textit{\textcolor{grey}{43.56}} &
\textit{\textcolor{grey}{35.29}} &
\textit{\textcolor{grey}{21.14}} &
\textit{\textcolor{grey}{28.73}} &
\textit{\textcolor{grey}{23.36}} &
\textit{\textcolor{grey}{26.63}} &
\textit{\textcolor{grey}{74.00}} &
\textit{\textcolor{grey}{90.48}} &
\textit{\textcolor{grey}{42.36}} &
\textit{\textcolor{grey}{4.80}} &
\textit{\textcolor{grey}{69.25}} &
\textit{\textcolor{grey}{57.03}} &
\textit{\textcolor{grey}{52.38}} &
\textit{\textcolor{grey}{41.65}} \\
\textbf{KNorm} & 10.00 & 7.98 & 11.05 & 11.90 & 8.92 & 4.83 & 21.24 & 18.39 & 23.43 & 40.00 & 51.87 & 17.14 &5.42  & 47.89 & 28.22 & 21.92 & 20.64 \\
\cellcolor{grey!10}~+ GraphKV & \cellcolor{grey!10}19.51 & \cellcolor{grey!10}16.02 & \cellcolor{grey!10}24.05 & \cellcolor{grey!10}25.45 & \cellcolor{grey!10}14.71 & \cellcolor{grey!10}11.01 & \cellcolor{grey!10}21.29 & \cellcolor{grey!10}20.29 & \cellcolor{grey!10}23.47 & \cellcolor{grey!10}43.50 & \cellcolor{grey!10}85.50 & \cellcolor{grey!10}31.52 & \cellcolor{grey!10}4.05 & \cellcolor{grey!10}61.60 & \cellcolor{grey!10}38.11 & \cellcolor{grey!10}36.00 & \cellcolor{grey!10}\textbf{29.76} \\
\textbf{SnapKV} & 23.45 & 23.37 & 37.14 & 42.60 & 34.60 & 20.08 &22.08 & 22.61 & 24.06 & 70.50 & 90.52 & 39.88 & 5.81 & 69.50 & 59.50 & 53.87 & 39.97 \\
\cellcolor{grey!10}~+ GraphKV & \cellcolor{grey!10}23.68 & \cellcolor{grey!10}25.37 & \cellcolor{grey!10}37.55 & \cellcolor{grey!10}43.21 & \cellcolor{grey!10}35.72 & \cellcolor{grey!10}20.94 & \cellcolor{grey!10}22.57 & \cellcolor{grey!10}22.86 & \cellcolor{grey!10}24.02 & \cellcolor{grey!10}70.50 & \cellcolor{grey!10}90.33 & \cellcolor{grey!10}39.48 & \cellcolor{grey!10}5.59 & \cellcolor{grey!10}69.25 & \cellcolor{grey!10}58.42 & \cellcolor{grey!10}56.10 & \cellcolor{grey!10}\textbf{40.35} \\
\textbf{PyramidKV} & 24.59 & 21.90 & 36.11 & 42.51 & 33.01 & 20.09 & 22.28 & 22.70 & 23.83 & 69.00 & 90.42 & 40.66 & 5.78 & 69.25 & 57.41 & 53.91 & 39.58 \\
\cellcolor{grey!10}~+ GraphKV & \cellcolor{grey!10}24.12 & \cellcolor{grey!10}23.57 & \cellcolor{grey!10}37.05 & \cellcolor{grey!10}44.13 & \cellcolor{grey!10}33.59 & \cellcolor{grey!10}20.67 & \cellcolor{grey!10}22.31 & \cellcolor{grey!10}22.87 & \cellcolor{grey!10}24.07 & \cellcolor{grey!10}70.00 & \cellcolor{grey!10}90.52 & \cellcolor{grey!10}40.09 & \cellcolor{grey!10}5.78 & \cellcolor{grey!10}69.12 & \cellcolor{grey!10}58.05 & \cellcolor{grey!10}55.46 & \cellcolor{grey!10}\textbf{40.09} \\

\arrayrulecolor{black}\midrule
\multicolumn{18}{c}{Mistral-7B-Instruct-v0.2, KV Size = 512} \\
\arrayrulecolor{black}\midrule
\textit{\textcolor{grey}{\textbf{Full}}} &
\textit{\textcolor{grey}{26.81}} &
\textit{\textcolor{grey}{33.19}} &
\textit{\textcolor{grey}{49.26}} &
\textit{\textcolor{grey}{43.02}} &
\textit{\textcolor{grey}{27.12}} &
\textit{\textcolor{grey}{18.78}} &
\textit{\textcolor{grey}{32.80}} &
\textit{\textcolor{grey}{24.16}} &
\textit{\textcolor{grey}{27.02}} &
\textit{\textcolor{grey}{71.00}} &
\textit{\textcolor{grey}{86.23}} &
\textit{\textcolor{grey}{42.64}} &
\textit{\textcolor{grey}{2.75}} &
\textit{\textcolor{grey}{86.98}} &
\textit{\textcolor{grey}{55.09}} &
\textit{\textcolor{grey}{53.01}} &
\textit{\textcolor{grey}{42.49}} \\
\textbf{KNorm} & 7.34 & 8.53 & 11.93 & 9.44 & 7.16 & 3.91 & 22.22 & 16.81 & 22.43 & 35.75 & 21.25 & 9.27 & 3.59  & 6.42 & 29.13 & 23.29 & 14.91 \\
\cellcolor{grey!10}~+ GraphKV & \cellcolor{grey!10}8.98& \cellcolor{grey!10}7.75 & \cellcolor{grey!10}12.50 & \cellcolor{grey!10}7.87 & \cellcolor{grey!10}7.83 & \cellcolor{grey!10}3.56 & \cellcolor{grey!10}21.58 & \cellcolor{grey!10}18.44 & \cellcolor{grey!10}22.66 & \cellcolor{grey!10}45.50 & \cellcolor{grey!10}26.79 & \cellcolor{grey!10}12.61 & \cellcolor{grey!10}4.24 & \cellcolor{grey!10}5.17 & \cellcolor{grey!10}32.22 & \cellcolor{grey!10}26.29 & \cellcolor{grey!10}\textbf{16.50} \\

\textbf{SnapKV}  & 24.71 & 27.92 & 49.01 & 38.39 & 25.02 & 17.53 & 23.69 & 23.59 & 24.61 & 67.00 &85.77 & 41.90 &2.81 &88.18 & 53.12 & 50.60& 40.24\\
\cellcolor{grey!10}~+ GraphKV & \cellcolor{grey!10}25.21 & \cellcolor{grey!10}29.21 & \cellcolor{grey!10}49.51 & \cellcolor{grey!10}39.15 & \cellcolor{grey!10}25.84 & \cellcolor{grey!10}18.15 & \cellcolor{grey!10}23.15 & \cellcolor{grey!10}23.75 & \cellcolor{grey!10}24.82 & \cellcolor{grey!10}67.00 & \cellcolor{grey!10}86.23 & \cellcolor{grey!10}43.32 & \cellcolor{grey!10}2.75 & \cellcolor{grey!10}88.36 & \cellcolor{grey!10}52.89 & \cellcolor{grey!10}51.23 & \cellcolor{grey!10}\textbf{40.60}\\

\textbf{PyramidKV}  & 24.25 & 26.40 & 48.48 & 40.22 & 24.97 & 17.62 & 23.18 & 23.33 & 24.40 & 67.50 &85.73 & 41.64 &3.03 &87.64 & 53.90 & 50.26 & 40.16\\
\cellcolor{grey!10}~+ GraphKV &\cellcolor{grey!10}24.62 & \cellcolor{grey!10}28.59 & \cellcolor{grey!10}49.36 & \cellcolor{grey!10}42.08 & \cellcolor{grey!10}25.83 & \cellcolor{grey!10}18.03 & \cellcolor{grey!10}23.16 & \cellcolor{grey!10}22.82 & \cellcolor{grey!10}24.07 & \cellcolor{grey!10}67.50 & \cellcolor{grey!10}85.20 & \cellcolor{grey!10}41.93 & \cellcolor{grey!10}2.57 & \cellcolor{grey!10}87.56 & \cellcolor{grey!10}54.18 & \cellcolor{grey!10}50.57 & \cellcolor{grey!10}\textbf{40.50} \\

\arrayrulecolor{black}\bottomrule
\end{tabular}
}
\caption{Performance comparison of GraphKV on LongBench.}
\label{table:longbench}
\end{table*}

\section{Experiment}
\subsection{Experimental Setup}

\subsubsection{Baseline Methods}
We integrate GraphKV with five state-of-the-art methods: \textbf{CAKE} \cite{qin2025cake} considers the temporal and spatial aspects of attention, \textbf{SnapKV} \cite{li2024snapkv} clusters recent attention, \textbf{PyramidKV} \cite{cai2024pyramidkv} employs a budget allocation strategy, 
\textbf{H2O} \cite{zhang2023h2o} uses cumulative attention and \textbf{KNorm} \cite{devoto2024simple} applies \(L_2\) norm of keys. For more detailed information, please refer to Appendix \ref{app:imp}.

\begin{figure*}[t!] 
  \centering
  \vspace{-0.4cm}\includegraphics[width=1\textwidth]{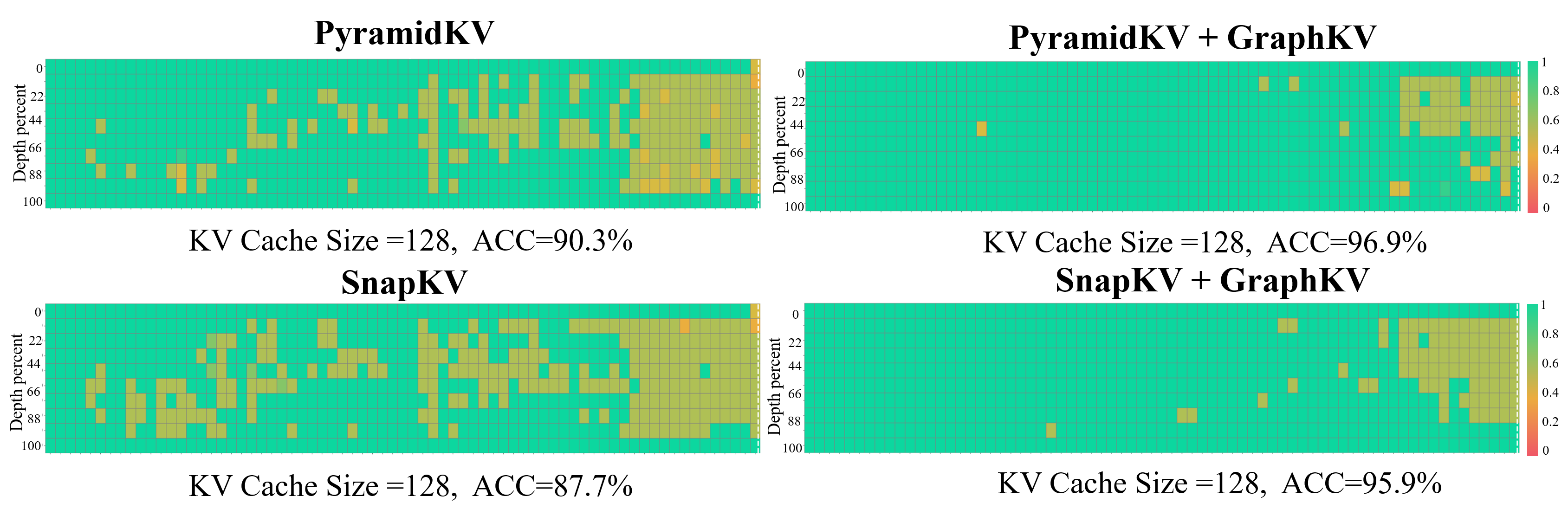} 
  \vspace{-1cm}
  \caption{Results of the Fact Retrieval Across Context Lengths (“Needle In A HayStack”) test in LlaMa-3-8B-Instruct with 8k context size in 128 KV cache size. The vertical axis of the table represents the depth percentage, and the horizontal axis represents the token length.} 
  \label{exp1}     
\end{figure*}

\subsection{Evaluation on LongBench}
The evaluation results for the LongBench dataset are presented in Table~\ref{table:longbench}. We evaluated three state-of-the-art models Llama2-7B-Chat, Llama3-8B-Instruct, and Mistral-7B-Instruct-v0.2, each configured with a fixed KV cache size of 512. As shown in Table~\ref{table:longbench}, a full KV cache yields the highest performance but is impractical for long-context applications due to its substantial GPU memory requirements. In contrast, our proposed GraphKV method achieves superior performance across all three models, surpassing other methods in the average score across 16 tasks. GraphKV shows a significant improvement on Retrieval-Based Passage (RB-P) dataset, evaluating the ability to understand cross-file dependencies, retrieve relevant code snippets, and generate accurate code completions in a multi-file programming context, while surpassing full-context model performance with only 10\% of the full budget. This demonstrates its effectiveness for long-context inference. Furthermore, integrating GraphKV with three token eviction strategies enhances performance, highlighting its flexibility and compatibility. Overall, GraphKV offers an effective and flexible solution for long-context inference, significantly reducing GPU memory usage while maintaining high performance.

\subsection{Evaluation on Needle In a Haystack}
In this experiment, we assess the Llama3-8B model under resource constraints, with a context length of 8,000 tokens and a KV cache size of 128. Figure \ref{exp1} shows retrieval accuracy on the Needle In a Haystack benchmark comparing PyramidKV, SnapKV, and Knorm with/without GraphKV integration. Notably, PyramidKV and SnapKV see notable improvements with GraphKV: PyramidKV's accuracy rises from 90.3\% to 96.9\% (+6.6\%), and SnapKV's from 87.7\% to 95.9\% (+8.2\%). These results highlight GraphKV's effectiveness in enhancing retrieval performance for long-context tasks under resource constraints.

\section{Ablation}
To further evaluate the design choices of GraphKV, we perform ablation studies to study the different components, including the choice of the number of source nodes, the number of adjacent nodes, and the number of propagation rounds, with experiments on some datasets in Longbench.

\subsection{Effect of Adaptive Source Nodes}

\begin{figure}[t]
    \centering
    \makebox[\columnwidth]{
        \includegraphics[width=1\columnwidth]{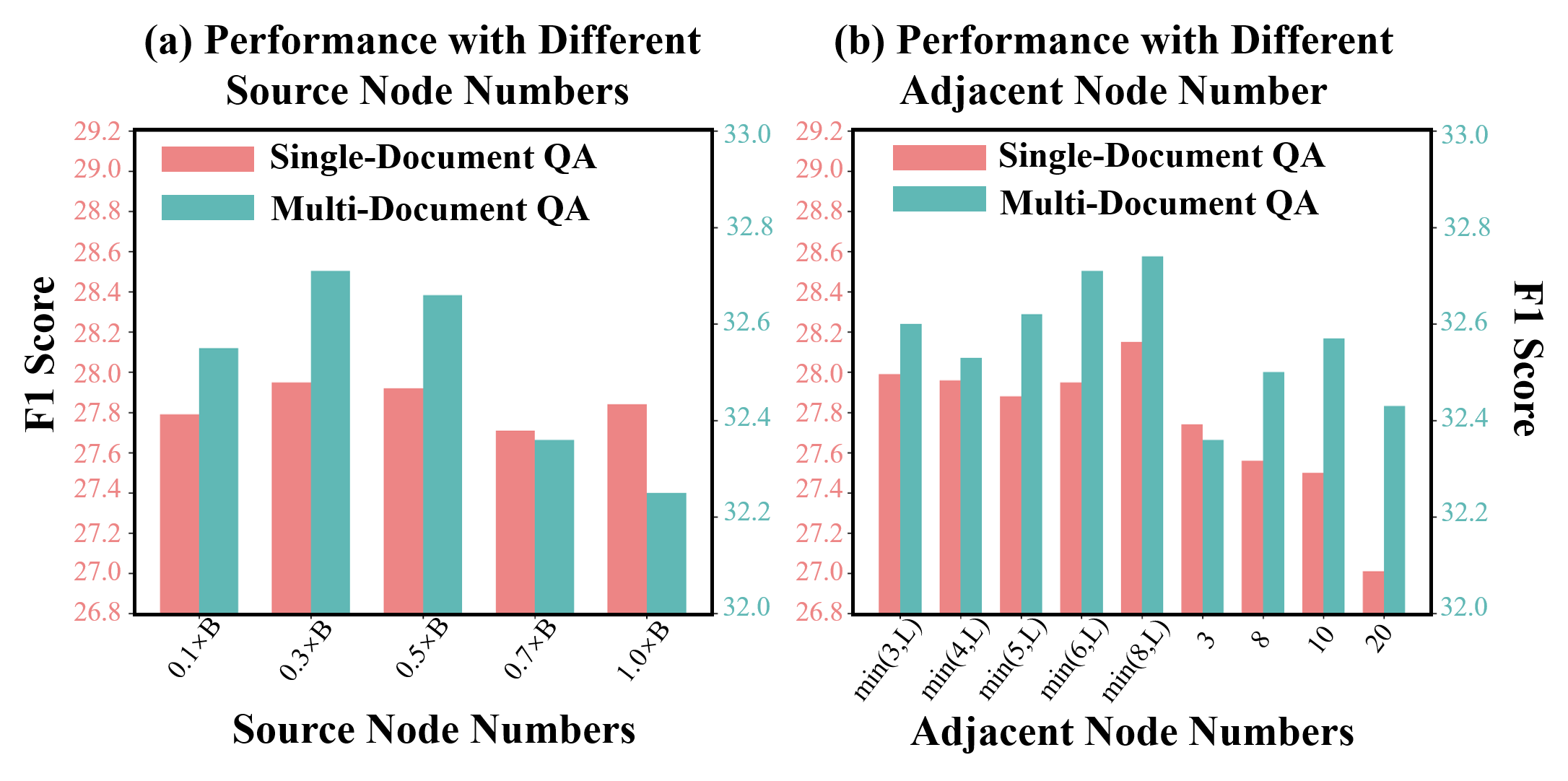} 
    }
    \caption{Impact of the number of adaptive source nodes and the number of adjacent nodes. ``B'' denotes the budget of the KV cache size. ``L'' denotes the length of input tokens $\times 10^{-3}$.}
    \label{fig:node}
\end{figure}

To evaluate the impact of the graph propagation range, we constructed sparse graphs with varying numbers of source nodes and conducted propagation tests. Specifically, we utilized a proportion of the KV cache budget as the number of source nodes. As shown in Figure \ref{fig:node} (a), a proportion of 0.3 $\times$ the KV cache budget achieved the best performance (0.3$\times$$B$). As the number of source nodes increased, the performance declined. This is because the number of source nodes is positively correlated with the number of nodes affected by the score updates, and a greater number of influenced nodes tends to introduce noise into the token importance distribution.

\subsection{Effect of Adjacent Nodes}
In addition to test the impact of the number of source nodes, we also explored the effect of the number of adjacent nodes for each source node. To adapt to query inputs of varying lengths, we set the number of adjacent nodes as the minimum between a predefined value and the query length. As shown in Figure \ref{fig:node} (b), the adaptive approach to determining the number of adjacent nodes generally outperforms a fixed setting. Under a fixed setting, increasing the number of adjacent nodes leads to poorer performance, which further indicates that a greater number of influenced nodes introduces noise into the token importance distribution.

\subsection{Effect of Propagation Round}

To evaluate the impact of propagation rounds on GraphKV, we integrated it with state-of-the-art methods, including CAKE, PyramidKV, SnapKV, and H2O. We conducted experiments under four settings: no propagation ($T$=0) and propagation rounds $T$=1, 2, and 3.

\begin{figure}[t] 
    \centering
    \makebox[\columnwidth]{
        \includegraphics[width=1\columnwidth]{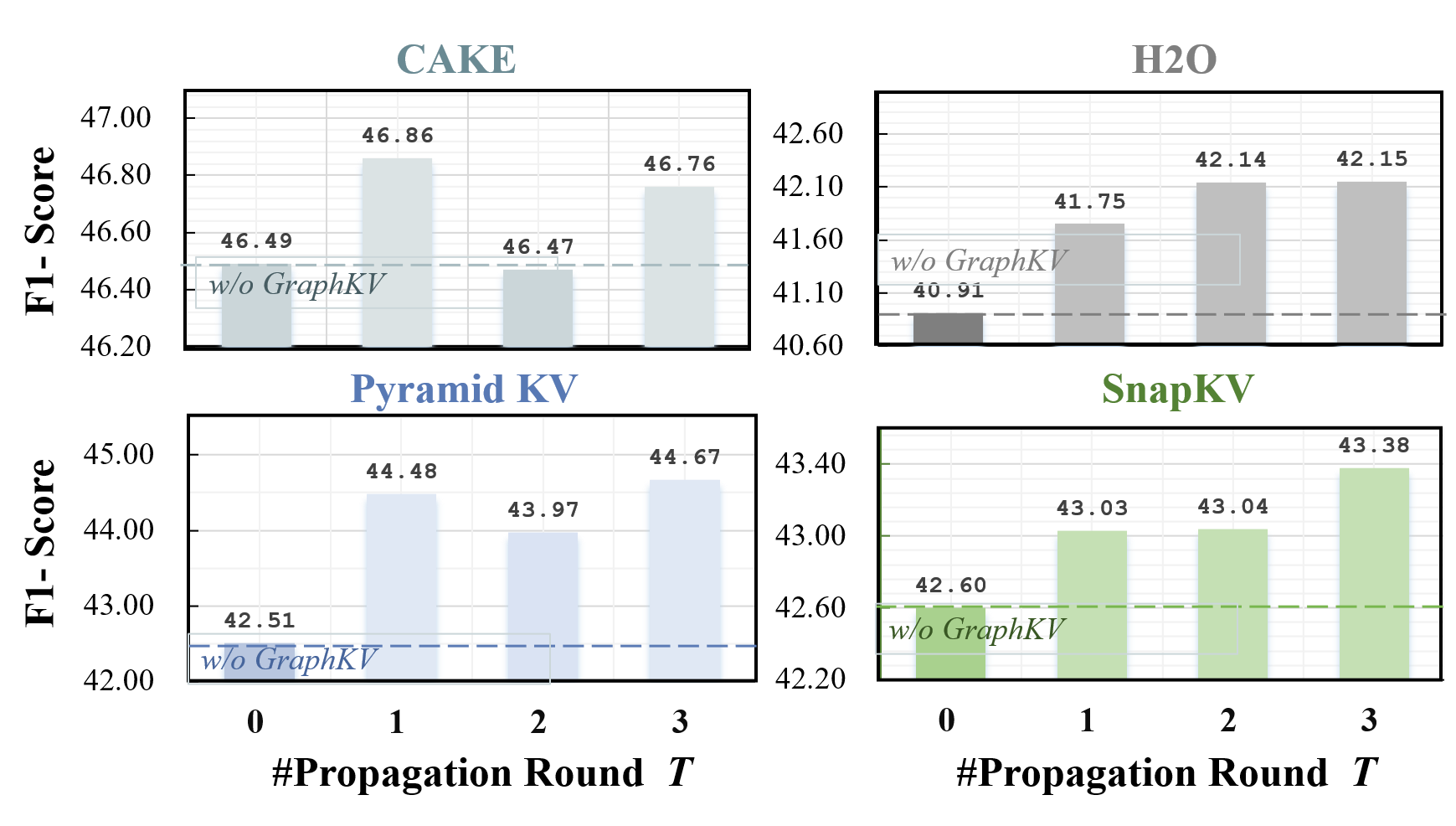} 
    }
    \caption{Impact on  the number of propagation rounds of GraphKV with decay signal propagation.}
    \label{fig:accuracy}
\end{figure}

 As shown in Figure \ref{fig:accuracy}, comparing $T$=0 and $T$=1, all four methods exhibited the most significant performance improvements after the first round of decay signal propagation. Notably, PyramidKV exhibited a significant performance, jumping from 42.51 to 44.48, surpassing FullKV after just one round. However, CAKE and PyramidKV showed slight performance dips at $T$=2, though still outperforming the non-propagated baseline. These results indicate that while performance fluctuates slightly with increased propagation rounds, overall, propagation substantially enhances performance.

\begin{figure*}[t]
    \centering
    \includegraphics[width=1\textwidth]{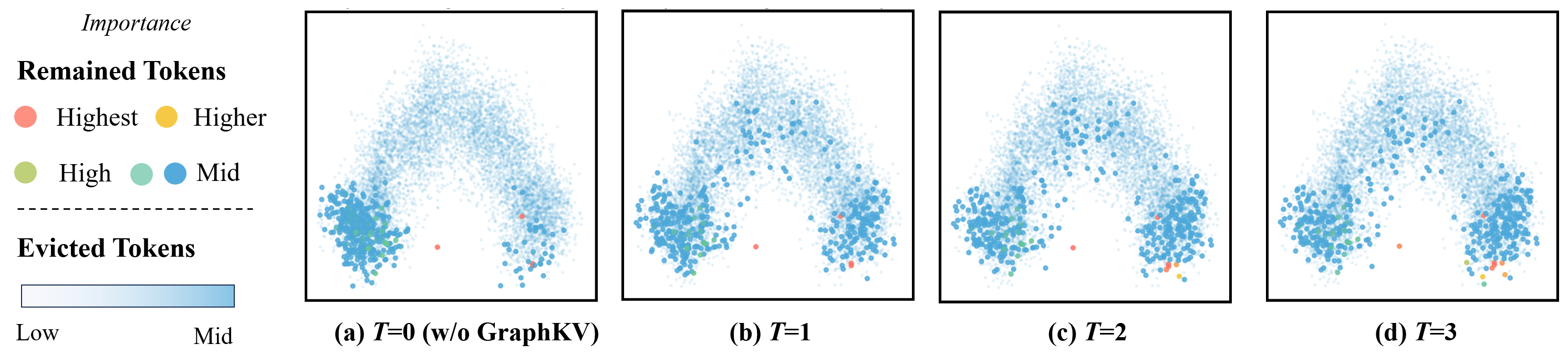}
    \hspace*{-18pt} 
    \caption{PCA visualizations in two dimensions of Keys with normalized importance scores for GraphKV across propagation rounds, where key vectors are colored by using a high-contrast colormap (sky blue to red), with retained tokens highlighted as larger, opaque markers.}
    \label{fig:pca}
\end{figure*}

As Figure \ref{fig:pca} shows, subfigures (a)–(d) show key vector distributions projected onto PCA components at steps $T$=0, 1, 2, and 3, respectively. Vectors are colored by normalized importance scores using a high-contrast colormap, with top5\% high-score tokens highlighted as larger, opaque markers. At $T$=0, key vectors of tokens with high importance scores are concentrated in a single region. After the first round of propagation, the token distribution becomes noticeably sparser. This change leads to significant performance improvements, validating that propagation reduces token redundancy.

\begin{figure*}[t]
    \centering
    \includegraphics[width=1\textwidth]{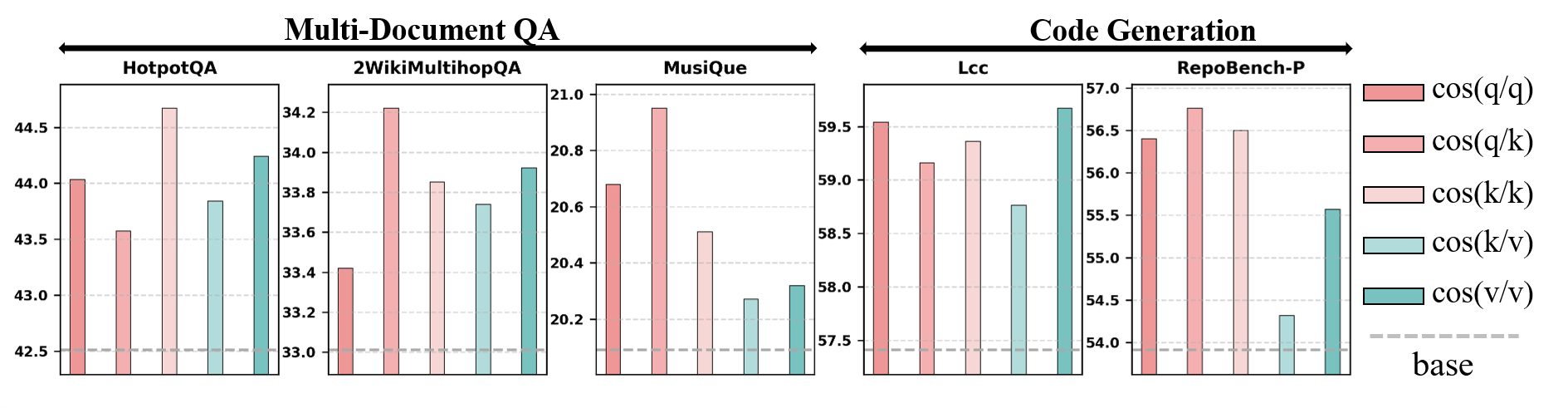}
    \vspace{-18pt}
    \caption{Performance comparison of GraphKV across datasets with different similarity graph edges.}
    \label{Different Graph Vertex}
     \vspace{-0.3cm}
\end{figure*}

\begin{table}[h]
    \centering
    \caption{Performance (f1 Scores or accuracy) comparison over three propagation signal types with 128 KV cache size across various datasets. The best result is highlighted in \textbf{bold}, the second best in \underline{underline}.}
    \label{tab:signal_performance}
    \resizebox{\columnwidth}{!}{%
    \begin{tabular}{lcccc}
        \toprule
        \textbf{Dataset} & \textbf{Decay ($-$)} & \textbf{Enhanced ($+$)} & \textbf{Evicted ($-\infty$)} & \textbf{Baseline} \\
        \midrule
        NarrativeQA      & \textbf{22.21} & \underline{21.71} & 5.09  & 21.35 \\
        Qasper           & \textbf{13.59} & 12.53 & 3.89  & \underline{13.46} \\
        HotpotQA         & \textbf{41.14} & 39.75 & 11.84 & \underline{40.68} \\
        2WikiMultihopQA  & \textbf{29.43} & 28.83 & 14.17 & \underline{28.94} \\
        MusiQue          & \underline{19.68} & \textbf{19.78} & 3.71  & 18.84 \\
        TREC             & \textbf{55.00} & \underline{52.50} & 41.00 & 51.50 \\
        SAMSum           & \textbf{39.01} & \underline{38.16} & 9.08  & 38.13 \\
        PassageCount     & \underline{68.92} & \textbf{69.50} & 65.29 & \textbf{69.50} \\
        Lcc              & \textbf{55.43} & \underline{55.04} & 8.33  & 54.49 \\
        RepoBench-P      & \textbf{53.94} & \underline{52.84} & 9.89  & 52.78 \\
        \midrule
        Average              & \textbf{39.84} & \underline{38.67} & 18.80 & 38.61 \\
        \bottomrule
    \end{tabular}%
    }
\end{table}
\section{Discussion}
Inspired by the relationship between key similarity and token's importance score, we propose a method for propagating decay signals based on graph structures. In the following sections, we further discuss this graph-based score updating paradigm.
\subsection{Exploration of similarity for Graph Edge}
In addition to key-to-key cosine similarity, we explore using query-to-key, query-to-query, key-to-value, and value-to-value similarities as decay signals for graph propagation. We assess their performance in multi-document QA and code generation tasks. Figure \ref{Different Graph Vertex} shows that these additional measures improve performance across various sub-datasets when used as edge-based decay signals. This underscores the effectiveness of avoiding storing too many highly similar tokens in GraphKV, supporting previous observation and motivation.


\begin{table}[t]
    \centering
    \small
    \begin{tabular}{ll}
    \toprule
    \textbf{Method} & \textbf{Decoding Latency (s)} \\
    \midrule
    \textbf{PyramidKV} & 5.133 \\
    
    PyramidKV+ GraphKV & 5.265 (+2.5\%) \\
    \textbf{SnapKV} & 5.474 \\
    SnapKV+ GraphKV & 4.912 (-10.2\%) \\
    \textbf{Knorm} & 6.031 \\
    Knorm+ GraphKV & 5.095 (-15.5\%) \\
    \bottomrule
    \end{tabular}
    \caption{The decoding latency of different methods on QMSum in LongBench at the KV budget of 512.}
    \vspace{-0.4cm} \label{tab:decoding_latency}
\end{table}
\subsection{Diversity of Graph Signal}

The experiments presented above validate the effectiveness of decay signal propagation in GraphKV. To comprehensively analyze graph signal propagation behavior, we further investigated different signal propagation modes. Specifically, we configured three signal types: decay signal ($-$), enhanced signal ($+$), and evicted signal ($-\infty$).

As presented in Table \ref{tab:signal_performance}, the decay signal propagation yields the most significant improvement over the baseline method, followed by the enhanced signal, which provides a marginal improvement of only 0.06, rendering it negligible. This further validates the critical role of GraphKV in leveraging decay signals to eliminate token semantic redundancy. Importantly, propagating the evicted signal leads to a substantial performance decline. This is primarily due to the naive and coarse approach of evicting all nodes connected to the source node, which unintentionally removes too many important tokens. 
 indicating a balance between importance and diversity should be achieved.

\subsection{Efficiency Analysis}

Table~\ref{tab:decoding_latency} gives the latency of GraphKV over three previous methods, demonstrating that GraphKV does not increase the inference latency, and even sometimes reduces it. This may be caused by that GraphKV increases model accuracy and hence prevents the model from over-long generation.

\section{Conclusion}
This paper investigates the critical limitations of static-importance-score-driven token eviction strategies in LLMs, which often fail to capture the relation between different tokens and suffer from duplicate KV cache. To address these challenges, we propose GraphKV, which models tokens as nodes and the similarity as edges. GraphKV dynamically refines token importance scores through an iterative decay-score-propagation mechanism, prioritizing contextually diverse and non-redundant tokens, leading to better performance in long-context scenarios.

\section{Limitation}
Although GraphKV substantially advances KV cache compression for long-context large language model inference through its innovative graph-based decay-signal-propagation framework for dynamic token relationship modeling, several limitations persist. The propagation mechanism depends on empirically determined hyperparameters, such as decay strength and propagation rounds, lacking a rigorous theoretical foundation for optimal configuration across diverse context complexities. Furthermore, due to computational constraints, evaluations are confined to 8B models, leaving the scalability and effectiveness of GraphKV on larger models unexplored. These gaps highlight the need for deeper theoretical analysis and broader empirical validation in future work.

\bibliography{main}

\appendix

\section{Appendix}
\subsection{Implementation Details}
\label{app:imp}
\noindent \textbf{Model Configuration.}
Our experiments include three state-of-the-art open-source LLMs. Specifically Llama2-7B-Chat \cite{touvron2023llama}, Llama-3-8B-Instruct \cite{grattafiori2024llama} and Mistral-7B-Instruct-v0.2 \cite{jiang2024identifying}. All experiments are conducted on a single NVIDIA H20 GPU. We use beam search according to ~\cite{cai2024pyramidkv, li2024snapkv}.

\noindent \textbf{Evaluation Tasks.}
To assess GraphKV's performance, we use two widely used benchmarks LongBench \cite{longbench} and Needle In a Haystack \cite{li2024snapkv}. LongBench is a bilingual, multi-task benchmark suite for LLMs, providing a comprehensive stress test for long prompt inputs. It consists of a diverse set of tasks, including question answering, summarization, reading comprehension, and code-related tasks, with input contexts ranging from a few thousand to over 100,000 tokens. The dataset spans multiple languages and domains, such as scientific literature, news, and dialogues, ensuring robust evaluation across varied scenarios. A detailed breakdown of the sub-tasks and their corresponding metrics is provided in Table~\ref{tab:longbench_subtasks} below. Needle In a Haystack assesses retrieval and reasoning via three components: Single-Needle Retrieval, Multi-Needle Retrieval, and Multi-Needle Reasoning, testing performance in complex contextual environments.

\begin{table}[ht!]
\centering
\small 
\scalebox{0.85}{
\begin{tabular}{lccc}
\toprule
Dataset & Avg len & Metric & \#data \\
\midrule
NarrativeQA & 18,409 & F1 & 200 \\
Qasper & 3,619 & F1 & 200 \\
MultiFieldQA-en & 4,559 & F1 & 150 \\
\midrule
HotpotQA & 9,151 & F1 & 200 \\
2WikiMultihopQA & 4,887 & F1 & 200 \\
MuSiQue & 11,214 & F1 & 200 \\
\midrule
GovReport & 8,734 & Rouge-L & 200 \\
QMSum & 10,614 & Rouge-L & 200 \\
MultiNews & 2,113 & Rouge-L & 200 \\
\midrule
TREC & 5,177 & Accuracy (CLS) & 200 \\
TriviaQA & 8,209 & F1 & 200 \\
SAMSum & 6,258 & Rouge-L & 200 \\
\midrule
PassageCount & 11,141 & Accuracy (EM) & 200 \\
PassageRetrieval-en & 9,289 & Accuracy (EM) & 200 \\
\midrule
LCC & 1,235 & Edit Sim & 500 \\
RepoBench-P & 4,206 & Edit Sim & 500 \\
\bottomrule
\end{tabular}
}
\caption{The dataset statistics in LongBench include several key metrics. The 'Avg len' (average length) is measured by the number of words for datasets in English (or code). 'Accuracy (CLS)' represents classification accuracy, while 'Accuracy (EM)' denotes exact match accuracy.}
\label{tab:longbench_subtasks}
\end{table}

\subsection{More Related Works}
\noindent\textbf{Graph-Based Methods.} Graph-based modeling \cite{scarselli2008graph} has emerged as a powerful paradigm for capturing complex relationships in structured data, with applications ranging from social networks \cite{fan2019graph} to natural language processing \cite{wu2023graph}. Graph Neural Networks (GNNs), leverage message-passing mechanisms to dynamically update node representations by aggregating information from adjacent nodes. This iterative process effectively captures dependencies that evolve over time or context, making it well-suited for tasks requiring relational reasoning. In the context of language modeling, graph structures have been used to represent syntactic dependencies or semantic relationships in text. Inspired by these capabilities, our GraphKV method formulates the KV cache as a graph with tokens as nodes and semantic similarities as edges. To the best of our knowledge, our approach is the first method to explore the application of graph-based methods to KV cache eviction in LLMs. 

\begin{table}[t]
    \centering
    \small
    \begin{tabular}{cc}
    \toprule
    \textbf{KV Cache Size (Tokens)} & \textbf{KV Cache Memory (GB)} \\
    \midrule
    128              & 0.016          \\
    256              & 0.031          \\
    512              & 0.063          \\
    1024             & 0.125          \\
    2048             & 0.250          \\
    16K              & 1.953          \\
    32K              & 3.906          \\
    64K              & 7.813          \\
    128K             & 15.625          \\
    \bottomrule
    \end{tabular}
    \caption{Memory consumption of KV cache at different context lengths for a 7B-parameter model with 32 attention heads and 128-dimensional key/value vectors. Memory calculations assume BF16 precision.}
    \label{tab:kv_memory}
\end{table}

\begin{table}[t]
    \centering
    \small
    \setlength{\tabcolsep}{3pt} 
    \resizebox{\columnwidth}{!}{ 
    \begin{tabular}{lccccc}
    \toprule
    \multirow{2}{*}{\textbf{Task}} & \multicolumn{5}{c}{\textbf{\# Source Nodes}} \\
    \cmidrule(lr){2-6}
    & 10\% & 30\% & 50\% & 70\% & 100\% \\
    \midrule
    \multicolumn{6}{c}{\textbf{KV Cache Size = 512}} \\
    \midrule
    Qasper & 24.86 & 24.47 & 24.45 & \textbf{25.20} & 24.53 \\
    NarrQA & \textbf{23.06} & 23.03 & 23.01 & 22.65 & 22.18 \\
    MFQA-en & 35.45 & 36.34 & 36.31 & 35.27 & \textbf{36.82} \\
    HotpotQA & \textbf{43.91} & 43.75 & 43.70 & 43.10 & 42.66 \\
    2WikiQA & 33.09 & \textbf{34.05} & 34.00 & 33.70 & 33.51 \\
    MusiQue & \textbf{20.65} & 20.33 & 20.29 & 20.28 & 20.57 \\
    \bottomrule
    \end{tabular}%
    }
    \caption{Performance of GraphKV with varying numbers of source nodes (as a percentage of the KV cache size, 512) across multiple tasks. Bold values indicate the best performance for each task. Abbreviations: Qas (Qasper), NarrQA (NarrativeQA), MFQA-en (MultiFieldQA-en), 2WikiQA (2WikiMultihopQA).}
    \label{tab:num_nodes}
\end{table}
\begin{algorithm}
\caption{Sparse Graph Construction}
\label{alg:sparse_graph}
\begin{algorithmic}[1]
\State \textbf{Input:} Token sequence $X=\{x_1,\dots,x_n\}$, scores $S=\{s_1,\dots,s_n\}$, $k$
\State Initialize nodes $O=\{o_1,\dots,o_n\}$
\State Initialize empty edge set $E = \emptyset$
\State $O_{\text{source}} \gets \{o_i \mid s_i \text{ in top-}k\}$ 
\For{$o_i \in O_{\text{source}}$}
    \For{$o_j \in O \setminus \{o_i\}$} 
        \State $E \gets E \cup \left\{\frac{\langle k_i,k_j \rangle}{\|k_i\|\|k_j\|}\right\}$
    \EndFor
\EndFor
\State \textbf{Output:} Sparse graph $G=(O,E)$
\end{algorithmic}
\end{algorithm}
\begin{algorithm}
\caption{Decay Signal Propagation}
\label{alg:decay_propagation}
\begin{algorithmic}[1]
\State \textbf{Input:} Sparse graph $G=(O,E)$, scores $\{s_1,\dots,s_n\}$, $T$, $m$
\State Initialize $s_j^{(0)} \gets s_j$ for all $o_j \in O$
\For{round $t = 1$ \textbf{to} $T$}
    \For{each source node $o_i \in O_{\text{source}}$}
        \State $N(o_i) \gets \{o_j \mid e_{ij} \text{ is top-}m \text{ among } \{e_{ik}\}_{k \neq i}\}$ 
        \For{each $o_j \in N(o_i)$}
            \State $s_j^{(t)} \gets s_j^{(t-1)} \cdot (1 - \cdot e_{ij})$ 
        \EndFor
    \EndFor
\EndFor
\State \textbf{Output:} Refined scores $\{s_1^{(T)},\dots,s_n^{(T)}\}$
\end{algorithmic}
\end{algorithm}

\label{sec:appendix}

\end{document}